\documentclass[10pt,twocolumn,letterpaper]{article}
\usepackage{3dv}
\makeatletter
\@namedef{ver@everyshi.sty}{}
\makeatother
\usepackage{times}
\usepackage{epsfig}
\usepackage{graphicx}
\usepackage{amsmath}
\usepackage{amssymb}
\usepackage{xcolor}
\usepackage{multirow}
\usepackage{multicol}
\usepackage{pythonhighlight}
\usepackage[utf8]{inputenc}
\usepackage{caption}
\usepackage{booktabs}
\usepackage{svg}
\usepackage[T1]{fontenc}
\usepackage{subcaption}
\usepackage{balance}
\usepackage{newfloat}
\usepackage{placeins}
\usepackage{enumitem}
\usepackage{tikz}
\usetikzlibrary{positioning}

\usepackage[pagebackref=true,breaklinks=true,letterpaper=true,colorlinks,bookmarks=false]{hyperref}

\newcommand\blfootnote[1]{%
  \begingroup
  \renewcommand\thefootnote{}\footnote{#1}%
  \addtocounter{footnote}{-1}%
  \endgroup
}

\setlist{topsep=4pt,itemsep=4.5pt}

\DeclareFloatingEnvironment[fileext=frm,placement={!ht},name=Listing]{listing}

\newcommand{\ARXIV}[2]{#1} 

\threedvfinalcopy 

\def\threedvPaperID{124} 

\graphicspath{ {./images/} }

\newcommand{\TPD}{{Torch-Points3D}}

\newcommand{\red}[1]{\textcolor{red}{#1}}
\newcommand{\blue}[1]{\textcolor{blue}{#1}}

\renewcommand*{\eqref}[1]{(\hyperref[#1]{\ref*{#1}})}
\newcommand*{\figref}[1]{Figure~\hyperref[#1]{\ref*{#1}}}
\newcommand*{\tabref}[1]{Table~\hyperref[#1]{\ref*{#1}}}
\newcommand*{\Subref}[1]{\hyperref[#1]{(\subref*{#1})}}
\newcommand*{\secref}[1]{Section~\hyperref[#1]{\ref*{#1}}}
\newcommand*{\coderef}[1]{Listing~\hyperref[#1]{\ref*{#1}}}

\ifthreedvfinal\fi
\begin{document}

\title{\TPD: A Modular Multi-Task Framework \\ for Reproducible Deep Learning on 3D Point Clouds}

\author{Thomas Chaton$^\star$\\
{\tt\small \qquad thomas.chaton.ai@gmail.com\qquad}
\and
Nicolas Chaulet$^\star$\\
Principia Labs\\
{\tt\small \qquad nicolas@principialabs.co.uk}\qquad
\and
Sofiane Horache\\
Centre de Robotique, CAOR \\
Mines ParisTech, PSL university\\
{\tt\small sofiane.horache@mines-paristech.fr}
\and
Loic Landrieu\\
LASTIG, Univ Gustave Eiffel, ENSG  \\
IGN, F-94160 Saint-Mande, France\\
{\tt\small loic.landrieu@ign.fr}
}

\maketitle

\begin{abstract}
We introduce \TPD, an open-source framework designed to facilitate the use of deep networks on 3D data. Its modular design, efficient implementation, and user-friendly interfaces make it a relevant tool for research and productization alike. Beyond multiple quality-of-life features, our goal is to standardize a higher level of transparency and reproducibility in 3D deep learning research, and to lower its barrier to entry.

In this paper, we present the design principles of \TPD , as well as extensive benchmarks of multiple state-of-the-art algorithms and inference schemes across several datasets and tasks.
The modularity of \TPD{} allows us to design fair and rigorous experimental protocols in which all methods are evaluated in the same conditions.

{The \TPD{} repository :  \url{https://github.com/nicolas-chaulet/torch-points3d}.}
\end{abstract}
\begin{figure}[ht!]
\centerline{\includegraphics[width=8.288cm,height=16.0cm,keepaspectratio]{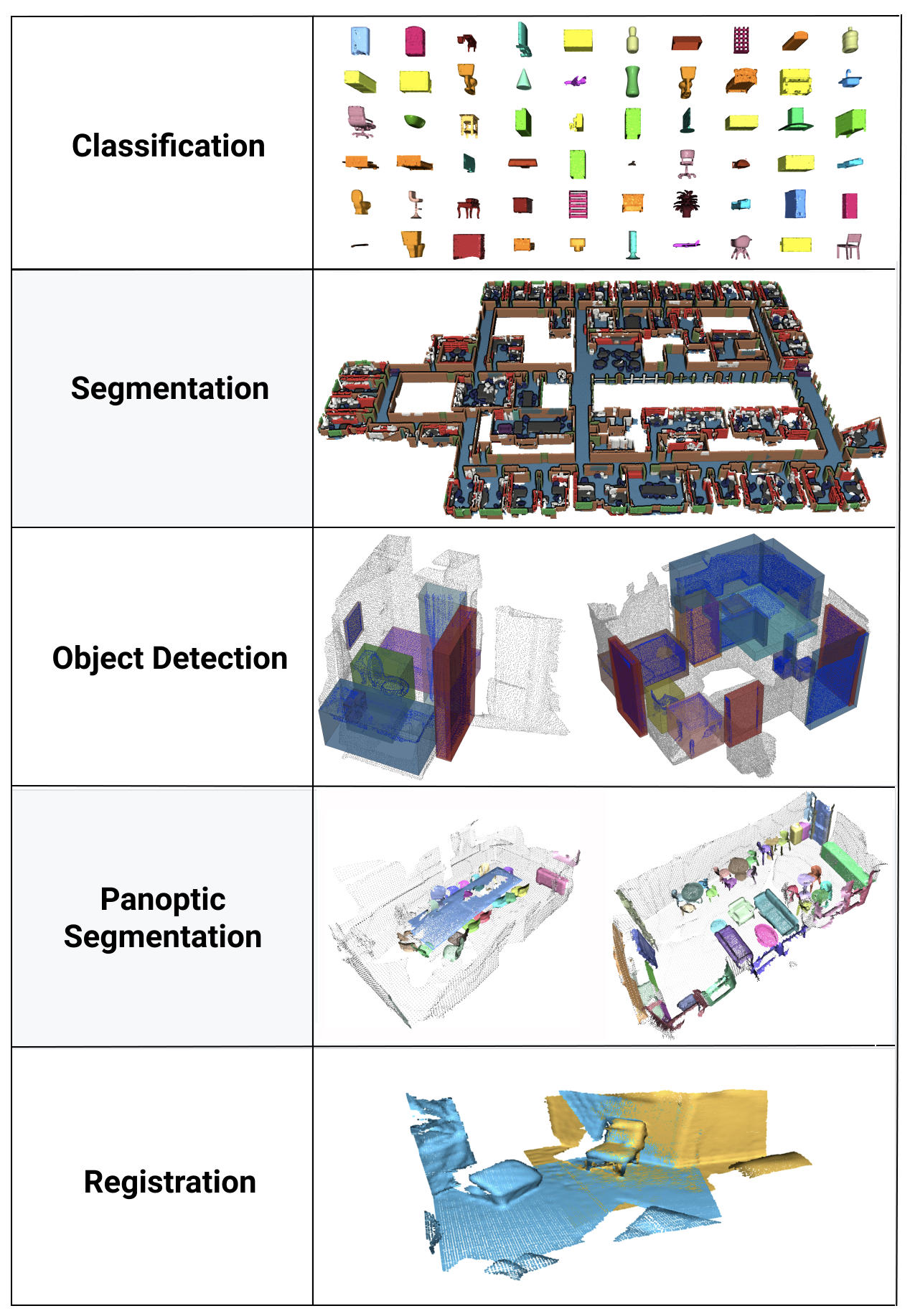}}
\caption{\TPD{} supports multiple tasks such as classification, segmentation, object detection, panoptic segmentation, and registration. All visuals have been produced by the framework.}
\label{fig:illustration}
\end{figure}
%
\section{Introduction}
\blfootnote{$^\star$ equal contribution}
In recent years, the field of automated analysis of 3D data has been transformed by the development of new dedicated neural network architectures.
This sudden spur in methodological advancements is reminiscent of the revolution undergone by image analysis in the early 2010s, initiated by AlexNet \cite{krizhevsky2012imagenet}.
The number of methods and papers dedicated to 3D data presented at major vision conferences is now on par with images, and keeps growing each year.

The young field of deep learning for 3D has greatly pushed forward the state-of-the-art performance of automated analysis of point clouds for numerous tasks. For example, the top performance on the indoor dataset S3DIS \cite{armeni20163d} (6-fold) have gone from $41.7$\% mIoU points (mean Intersection over Union) in 2017 \cite{qi2017pointnet}, to $62.1\%$ only a year later \cite{landrieu2018large}, and up to $70.6$\% in 2019 \cite{thomas2019kpconv}.
While this rapid methodological development is of course beneficial to the community, its fast pace comes with several shortcomings:
\begin{itemize}
\item Adding new datasets, tasks, or neural architectures to existing approaches requires a substantial commitment, often tantamount to a complete re-implementation. This limits the use of new networks, and prevents exhaustive comparisons.
\item Handling large 3D datasets efficiently requires a significant time investment, and overcoming many implementation pitfalls. This creates soft barriers to entry, restraining the diffusion of new ideas.
\item There is no standard approach for inference scheme and metrics in research papers. This makes assessing the intrinsic performance of new algorithms difficult, and their reproducibility not always straightforward.
\end{itemize}
%
In this paper, we introduce \TPD, a flexible and powerful development framework aiming to address these issues. In short, the purpose of our framework is to become for 3D point clouds what \href{https://github.com/pytorch/vision}{\texttt{torchvision}} \cite{marcel2010torchvision} or \href{https://github.com/rusty1s/pytorch_geometric}{\texttt{pytorch-geometric}} \cite{Fey2019Fast} have become for images and graphs respectively. More generally, our goal is to address the growing technical debt pervasive to machine learning research codes. This is particularly crucial for 3D data, for which many steps require special care in order to not invalidate investigations, from data loading and preprocessing to the computation of performance metrics. By proposing tried and tested implementations, which only get more robust as the user community grows, we aim to further increase the rigor of 3D deep learning.

\TPD \ is intended for novices as much as experts. It provides intuitive interfaces with most open-access 3D datasets, re-implementations of many of the top-performing networks, classic data augmentation schemes and validated performance metric. This allows researchers to focus on the development of core algorithms and test them on all available datasets with minimal effort. The different components of \TPD \ are highly customizable and can be plugged into one another with a unified system of configuration files. Users can then easily swap backbone networks for a given task, leading to the efficient selection of the best-suited algorithms, as well as facilitating comparison with new approaches. On this front, our framework makes it easy to standardize experimental protocols, ensuring both reproducibility and that models' performances are evaluated on equal footing.

Finally, we propose a multitude of quality-of-life features such as open logs with Weight and Biases \cite{wandb}, versatile model configuration handling with Hydra \cite{hydra}, and bespoke visualization functions as illustrated in \figref{fig:illustration}.

To illustrate the capabilities of \TPD, we propose several numerical experiments: 
\begin{itemize}
\item We evaluated the performance of different backbone networks in a recent object detection method.
\item We benchmark different methods over several datasets with a unified protocol, aiming to assess their inherent performance.
\item We quantify the benefit of implemented test-time enhancers, such as voting schemes.
\item We present our point clouds registration implementation within our framework, combining recent papers and reaching state-of-the-art performance.
\item We share key findings about speed enhancing procedures that can be leveraged on any model supported by the framework.
\end{itemize}
\section{Related Work}
The first deep learning methods for 3D point clouds analysis relied on image \cite{boulch2017unstructured} or voxel-based representations \cite{Riegler2017OctNet, tchapmi2017segcloud}.
PointNet \cite{qi2017pointnet, QiYSG17PointNetPP} was the first network whose architecture was specifically designed to handle unordered 3D point clouds. Since then, a multitude of approaches have been proposed, see the comprehensive review by Guo \etal \cite{guo2020deep}.
%
%
%
Manipulating large 3D point clouds requires extensive implementations and to this end, several open-source frameworks have been proposed.
\paragraph{\href{https://github.com/NVIDIAGameWorks/kaolin}{Kaolin}} Krishna Murthy J. \etal from Nvidia shared a Pytorch framework aiming to accelerate 3D deep learning research \cite{kaolin2019arxiv}. It implements boilerplate code for handling meshes, voxels, and point clouds.
\paragraph{\href{https://github.com/facebookresearch/pytorch3d}{Pytorch3D}} Nikhila Ravi \etal proposed another Pytorch-based framework, similar to Kaolin, for 3D computer vision research \cite{ravi2020pytorch3d}.
Its key features include bespoke data structure for storing and manipulating meshes, a differentiable mesh renderer, camera position optimization, bundle adjustment, and several mesh-based deep models \cite{gkioxari2019mesh}.
\paragraph{\href{https://github.com/poodarchu/Det3D}{Det3D}} Zhu Benjin. \etal open-sourced a 3D Object Detection Pytorch toolbox \cite{zhu2019class}, providing out-of-the-box
implementations of many 3D object detection algorithms  \cite{lang2019pointpillars, yang2018pixor}, as well as compatibility with several datasets such as KITTI \cite{geiger2013vision} and nuScenes \cite{caesar2020nuscenes}.
\paragraph{\href{https://github.com/open-mmlab/OpenPCDet}{OpenPCDet}} \!\!\!\!and {\href{https://github.com/open-mmlab/mmdetection3d}{\bf MMDetection3D}} \cite{mmdetection3d_2020} are open-source 3D object detection PyTorch toolboxes, part of the OpenMMLab project developed by CUHK Multimedia Lab.~\\

However, a unifying framework for multi-tasks, multi-models, multi-datasets for reproducible 3D point clouds deep learning has yet to be proposed. In this paper, we introduce \TPD, which aims to answer this need.

\section{The Framework}
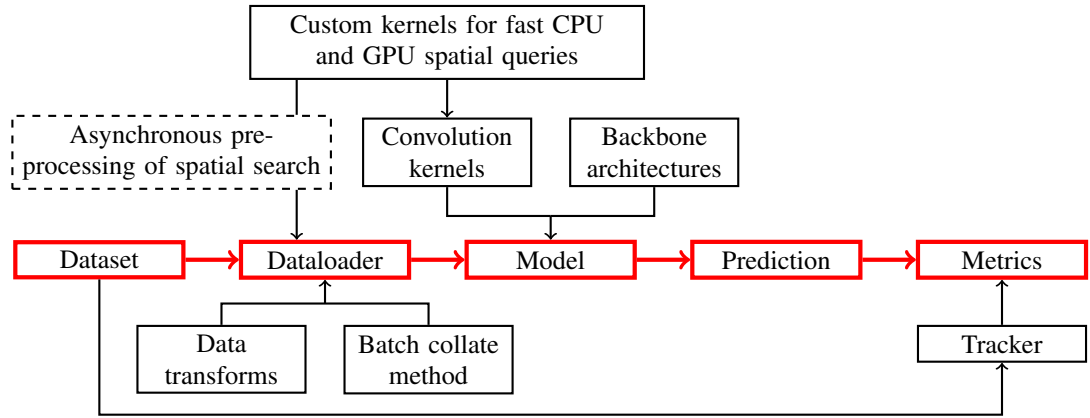
\begin{figure*}[!t]
\centering
\begin{tikzpicture}[
primary/.style={ultra thick, draw=red, text width=2cm, align=center},
secondary/.style={thick, draw=black,text width=2cm, align=center}]
\node[primary] at (0,0) (dataset) {Dataset};
\node[primary] at (3,0) (dataloader) {Dataloader};
\node[primary] at (6,0) (model) {Model};
\node[primary] at (9,0) (prediction) {Prediction};
\node[primary] at (12,0) (metrics) {Metrics};

\node[secondary, draw=black, below left = 6mm and -9mm of dataloader]  (transforms) {Data transforms};
\node[secondary, draw=black, below right = 6mm and -9mm of dataloader] (collate) {Batch collate
method};

\node[secondary, draw=black, above left = 7mm and -9mm of model]  (kernels) {Convolution
kernels};
\node[secondary, draw=black, above right = 7mm and -9mm of model] (backbone) {Backbone
architectures};

\node[secondary, draw=black, text width=5cm,  above = 5mm of kernels] (queries) {Custom kernels for fast CPU and GPU spatial queries};

\node[secondary, draw=black, text width=4cm, dashed,  left = 4 mm of kernels] (preprocessing) {Asynchronous pre-processing of
spatial search};

\node[secondary, draw=black, below = 6mm of metrics] (tracker) {Tracker};

\draw[ultra thick, red, ->] (dataset) -- (dataloader);
\draw[ultra thick, red, ->] (dataloader) -- (model);
\draw[ultra thick, red, ->] (model) -- (prediction);
\draw[ultra thick, red, ->] (prediction) -- (metrics);

\draw[thick, black, -] (transforms) |- ([yshift=-9pt] dataloader.south);
\draw[thick, black, -] (collate) |- ([yshift=-9pt] dataloader.south);
\draw[thick, black, ->] ([yshift=-9pt] dataloader.south) -- (dataloader);

\draw[thick, black, ->] (queries) -- (kernels);

\draw[thick, black, -] (kernels) |- ([yshift=+9pt] model.north);
\draw[thick, black, -] (backbone) |- ([yshift=+9pt] model.north);
\draw[thick, black, ->] ([yshift=+9pt] model.north) -- (model);

\coordinate (offset1) at ([xshift=-2cm] queries.south);
\draw[thick, black, -] (offset1) -- (offset1 |- preprocessing.north);
\draw[thick, black, ->] (offset1 |- preprocessing.south) -- (offset1 |- dataloader.north);

\coordinate (offset2) at ([yshift=-1.8cm] dataset.south);
\draw[thick, black, ->] (dataset) -- (offset2) -- (offset2 -| tracker.south) -- (tracker);
\draw[thick, black, ->] (tracker) -- (metrics);

\end{tikzpicture}
\caption{Overall architecture of the framework, with data flow  highlighted in red. \texttt{Dataset} implements the core loading mechanism of raw data and creates objects containing the points' positions, relevant features, and labels. Those objects are then passed through data augmentation transforms and aggregated into batches in the \texttt{Dataloader}. They are finally passed to the \texttt{Model}, which outputs the prediction. A tracker evaluates the performance from predefined \texttt{Metrics} and  publish results on the console and/or directly on Weight and Biases (\url{wandb.ai}).
}
\label{fig:pipeline}
\end{figure*}
\TPD{} was written from scratch according to the following design principles: it should be modular, extendible, and support multiple models, tasks, and datasets.
\figref{fig:pipeline} illustrates the different components of our framework and how they interact together. A key design principle is that the components are independent from one another allowing users to plug and play their own contributions. This could be a dataset, a custom convolution or a new data augmentation strategy for example.  In this section, we illustrate how these ideas translate into a versatile, easy-to-use interface.
%
\subsection{Dataset Handling}
The growing number of large-scale 3D public datasets has a beneficial effect on both the academic community and industrial actors interested in automated 3D point clouds analysis. While image formats have been standardized for years, this is not the case for 3D data. Hence, downloading, reading, cleaning, and processing data into a deep learning-ready format requires specific implementations, discouraging researchers to perform experiments on many datasets.

Building on  \href{https://github.com/rusty1s/pytorch_geometric}{\texttt{pytorch-geometric}} implementations, we propose an adapted interface for handling 3D datasets, from automatic downloading to data-augmentation.
To ensure maximum versatility, these operations can be set in a compact and modular configuration file system.
In \secref{sec:datasets}, we present the different datasets currently implemented in \TPD.
\begin{listing}
\caption{Configuration file \texttt{shapeNet-fixed.yaml} for the Shapenet dataset, with a fixed number of sampled points per object.}
\label{code:shapenet}
\begin{python}
data:
  class: shapenet.ShapeNetDataset
  task: segmentation # associated task
  dataroot: ./data #data path
  normal: True #use normals
  pre_transforms: # preprocessing
    - transform: NormalizeScale #to unit sphere
    - transform: GridSampling3D
      params: #size of voxel
        size: 0.02
  train_transforms: # Train data augmentation
    - transform: FixedPoints
      lparams: [2048] # random sampling
    - transform: RandomNoise 
      params: #Gaussian noise
        sigma: 0.01
        clip: 0.05
  test_transforms: #Test data augmentation
    - transform: FixedPoints
      lparams: [2048] # random sampling
\end{python}
\end{listing}
\subsection{Modular Model Configuration}
The majority of competitive deep learning networks for 3D analysis rely on the 3-step, U-net like approach initially proposed by PointNet++ \cite{QiYSG17PointNetPP}:
\begin{itemize}
    \item[(i)]\textbf{Encoding:} The input point cloud is iteratively subsampled, and local features are computed for each point of a subsampling level based on  neighboring points in the previous subsampling level. This step is built upon a \texttt{down\_conv} operation, which varies for different networks. 
   \item[(ii)]\textbf{InnerMost:} A global descriptor per instance is computed by pooling the last subsampling level into a single vector. This embedding can be processed further with fully connected layers. 
    \item[(iii)]\textbf{Decoding:} The learned features are interpolated back from lower sub-samplings levels, up to the original point clouds, and processed based on their neighbors' features, forming the \texttt{up\_conv} operation. Features at mirror-level of subsampling for the encoder can be concatenated for additional spatial precision, through so-called \texttt{skip} connections. The result of the innermost module can also be concatenated with point features at different subsampling level.
\end{itemize}
Based on this versatile architecture, we propose a system of configuration files able to encode most segmentation networks. For example, the official single-scale PointNet++ is implemented by the configuration file in \coderef{code:ptnpp}.
\begin{listing}
\caption{Configuration file for PointNet++ \cite{QiYSG17PointNetPP}.}
\label{code:ptnpp}
\begin{python}
  pointnet2:
    class: pointnet2.PointNet2 # Model class
    conv_type: "DENSE" # Convolution type
    down_conv: #encoder
      module_name: PointNetMSGDown
      npoint: [512, 128] #subsampling levels
      radii: [[0.2], [0.4]] #neigh. radius
      nsamples: [[64], [64]] #neigh. count
      down_conv_nn: [[[FEAT+3, 64, 64, 128]],
         [[128+3, 128, 128, 256]]]
    innermost: #process learned feature
        module_name: GlobalDenseBaseModule 
        nn: [256 + 3, 256, 512, 1024]
    up_conv: #decoder
        module_name: DenseFPModule
        up_conv_nn:
            [[1024 + 256, 256, 256],
              [256 + 128, 256, 128],
              [128 + FEAT, 128, 128, 128]]
        skip: True # use skip connection
    mlp_cls: #produce class scores
        nn: [128, 128]
        dropout: 0.5
\end{python}
\end{listing}
The model \texttt{pointnet2} described above can now be trained on any supported dataset such as Shapenet \cite{shapenet2015}   with a simple command:
\begin{python}
python train.py task=segmentation \
dataset=shapenet-fixed model_type=pointnet2 \
model_name=PointNet2
\end{python}

\subsection{Implemented Networks}
\TPD{} implements several convolution methods which present an interest in terms of performance or versatility: PointNet, RandLANet, KPConv, RS-CNN, and Minkowski Engine.
The PointNet-based architectures \cite{qi2017pointnet,QiYSG17PointNetPP} are the simplest point convolution methods, making them both easy to use and understand.
The convolution kernels for RandLANet \cite{hu2019randla} allows for efficient point clouds processing with a random sampling strategy. KPConv \cite{thomas2019kpconv} proposes a kernel-based generalization of 2D convolution to 3D point clouds, and RS-CNN \cite{liu2019rscnn} capture the complexity of local shapes by modeling spatial relationships between points. Minkowski Engine \cite{choy20194d} relies on a fine-grained voxelization of point clouds, efficiently processed with sparse-convolutions.

{These convolution schemes can be integrated into a backbone network architecture for semantic segmentation or object classification for example. We propose several such backbone, from a simple U-Net, to a multi-scale architecture, and more modern ResNet-like architectures as proposed in \cite{choy20194d} or \cite{thomas2019kpconv}.}

{Finally, the framework also implements  task-specific heads for object detection and Panoptic segmentation. In aprticular, VoteNet \cite{qi2019deep} uses Hough voting to regress bounding boxes on 3D point clouds, and  PointGroup \cite{jiang2020pointgroup} uses a clustering scheme to perform instance segmentation.}

While the compact configuration format described in the last section is designed to make U-Net-like networks easier, users can also easily define their own types of configuration file and model architectures in \TPD{}. For example, VoteNet and PointGroup use custom configurations. Our configuration system is meant to be flexible, and existing configurations serve more as guides for newcomers than rigid templates.
At the time of writing, superpoint-based \cite{landrieu2018large, landrieu2019point} and multi-modal methods (eg. 3D points + images \cite{dai20183dmv}) are not yet implemented. However, we plan to add both in the near-future.
\subsection{Multi-Task Support}
The ability of our framework to handle different tasks ensures its versatility and allows multi-source supervision  \cite{lin2013holistic}. We have currently implemented five tasks, illustrated in \figref{fig:illustration}, and their associated losses and metrics: classification, semantic segmentation, panoptic segmentation \cite{kirillov2019panoptic}, registration, and object detection.

Adding new tasks with their associated datasets and metrics can be done in isolation from the rest of the project, allowing users to extend the framework without necessarily understanding its inner working in details.
\subsection{Transparency and Reproducibility}
Reproducibility of experiments is not only necessary when assessing the suitability of different networks to a given task or dataset, but also to back scientific claims in academic papers. To this end, we have ensured the integrated compatibility of our framework with the Hydra configuration system \cite{hydra} as well as the experiments tracker \href{https://github.com/wandb/client}{Weight and Biases} \cite{wandb} (\url{wandb.ai}).
This online tool can store training runs along with their logs, metric 
visualization, configuration files, git commit hash, and our custom \href {https://docs.python.org/3/library/pickle.html}{Pickle}-based checkpoints. This total transparency allows users to compare their own experiments with our reference runs, and the models can be directly downloaded for fine-tuning tasks. In 
the Appendix, we reports an example of log visualization hosted on \url{wandb.ai.}

Another benefit of our unified approach is standardizing the learning and testing procedure. Indeed, the field of 3D analysis lacks a common ground when it comes to evaluation and augmentation strategies, both at test and training time. This makes experiments across different papers hard to compare, and could potentially obscur the intrinsic performance of new models. In \secref{sec:benchmark}, we propose standard protocols for different datasets and reproduce an array of experimental results.
\subsection{Supported Datasets}
\label{sec:datasets}
\TPD{} supports multiple academic datasets with automatic data download, pre-processing, as well as automatic result submission when available.
\begin{itemize}
    \item\textbf{\href{http://kaldir.vc.in.tum.de/scannet_benchmark/}{ScanNet}} is an indoor RGBD dataset containing $1\,201$ train scenes and $312$ test scenes \cite{dai2017scannet}. It can be used for semantic segmentation, object detection, and panoptic segmentation.
    \item\textbf{\href{http://buildingparser.stanford.edu/dataset.html}{S3DIS}} is a large-scale indoor RGB point cloud dataset covering three separate office buildings, over $6\,000\text{m}^2$, and containing $278$ million points with instance-level object and semantic annotations.
    We implement three different sampling for batch-training, based on rooms \cite{armeni20163d}, cubes \cite{QiYSG17PointNetPP}, or spheres \cite{thomas2019kpconv}.
    \item\textbf{\href{https://modelnet.cs.princeton.edu/}{ModelNet10/40}} is a dataset composed of over $12\,000$ CAD models from $10$ and $40$ categories \cite{wu20153d}.
    \item\textbf{\href{www.shapenet.org
}{Shapenet}} is a collection of over $200\,000$ CAD models annotated across a hierarchy of $3\,135$ classes \cite{shapenet2015}. On top of classification and semantic segmentation, \TPD{} also implements the task of hierarchical semantic segmentation, as well as adapted metrics.
\item\textbf{\href{https://3dmatch.cs.princeton.edu/}{3DMatch}} is an RGBD dataset \cite{zeng20163dmatch} widely used for 3D reconstruction and point cloud registration.
\item\textbf{\href{http://www.cvlibs.net/datasets/kitti/eval_odometry.php}{KITTI Odometry}} contains $21$ sequences of LiDAR frames \cite{Geiger2012CVPR}, with ground truth poses for the first ten. KITTI Odometry is commonly used as benchmmark for SLAM LIDAR, but  can also be used to train and evaluate point cloud registration networks (\cite{choy2019fully}).
\end{itemize}

\subsection{Built-in Visualization}
\begin{figure}[]
\centerline{\includegraphics[scale=0.2]{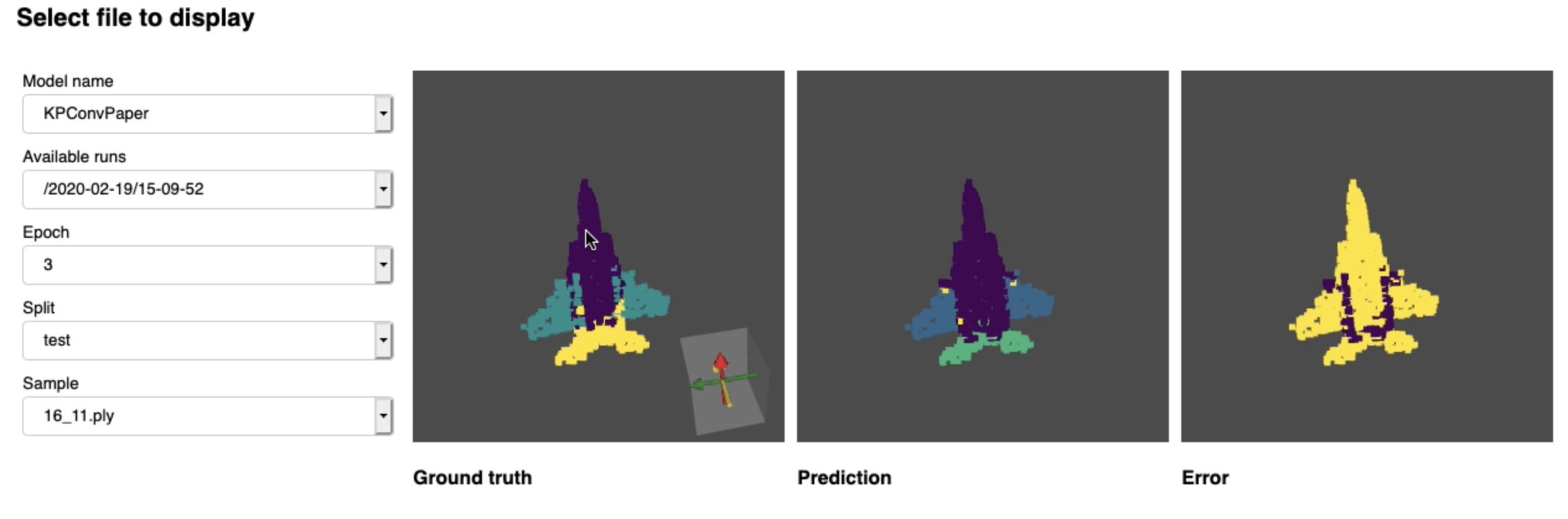}}
\centerline{\includegraphics[scale=0.15]{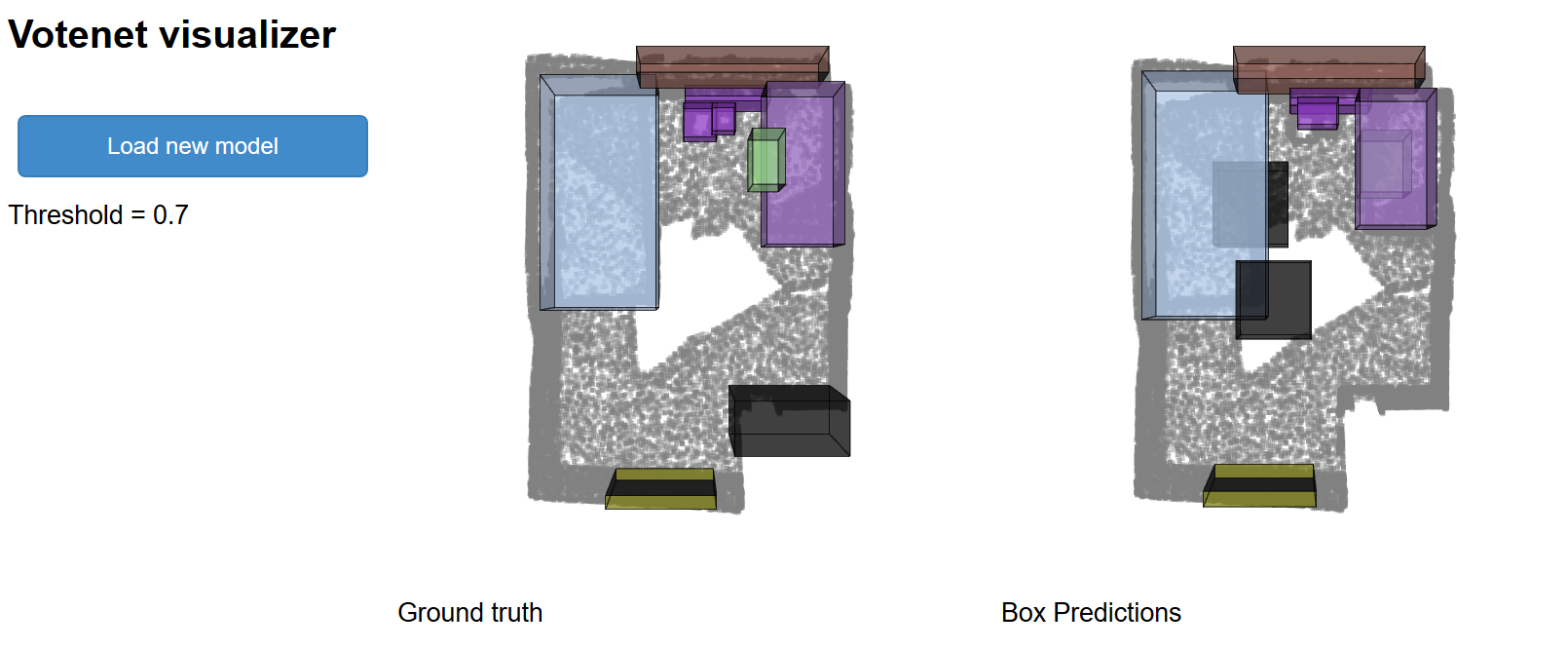}}
\centerline{\includegraphics[scale=0.2]{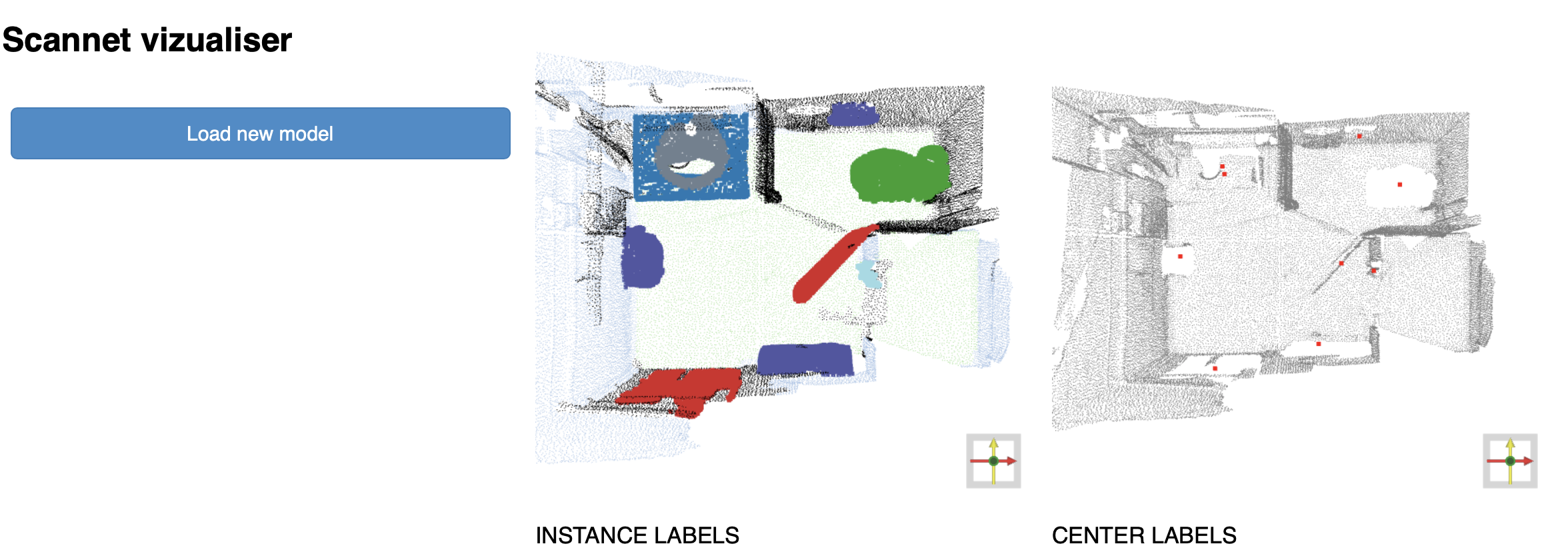}}
\caption{Several of the visualization tools available.}
\label{fig:viz}
\end{figure}

\TPD{} provides several custom visualization tools directly available within notebooks and using the dashboarding library \href{https://github.com/holoviz/panel}{panel}. This feature can be used to explore datasets, debug models, or illustrate predictions, as shown in \figref{fig:viz}.
\subsection{Ease-of-Use}
{As corroborated by the numerous pull-requests submitted by contributors, and testimonies from industrial partners, \TPD{} is accessible despite the quantity of code.} {Adding new datasets, new models or new base modules can be done without interfering with the rest of the code base while benefiting from existing methods for data augmentation, metric evaluation or training procedures at zero cost.} 
{A convenient way to start familiarizing one self with the framework, is by running the proposed illustrative IPython \href{https://github.com/nicolas-chaulet/torch-points3d/tree/master/notebooks}{notebooks}. More details as well as installation instructions are available in the Appendix.}
\if 1 0
\subsection{Collate Strategies}

Through community efforts to find efficient ways to process pointclouds, five different methods have appeared.

\paragraph{DENSE FORMAT}

Similar to image batching, multiple pointcloud instances can be batched on a new dimension. 
It creates a 3-d tensor $\mathbb{R}^{B,N,F}$ with B batch size, N number of points and F points dimension, typically 3 or 6 (with rgb)
Aggregation operation are fast, as well optimized on gpus.
However, a drawback appears when batching pointcloud instances with different sizes. Pytorch3D [] introduced the \textbf{PACKED Format} as a solution where instances are padded with shadow points (zeros features), which is sub-optimal in term of memory and speed.
Another solution, would be to randomly sample each instance to force the same number of points. However, forcing this condition can be damaging when instances differ largely in shape and forms.
This format was firstly introduced in the PointNet++ official implementation [] and re-used later on by RS-CNN and RandLaNet [].

\paragraph{HETEROGENEOUS FORMATS}

The \textbf{MESSAGE PASSING}, \textbf{PARTIAL DENSE}, \textbf{SPARSE} formats relies on multiple pointcloud instances being batched into an single instance by concatenating their feature matrices in the node dimension. 
In addition, an automatically generated assignment vector ensures that node-level information is not aggregated across instances, e.g., when executing global aggregation operators. 
However, those 3 formats differ in their way to perform neighborhood aggregation.

\[
\small
\left[
\begin{bmatrix}
{\red{x_1}} & {\red{y_1}} & {\red{z_1}} \\
{\red{x_2}} & {\red{y_2}} & {\red{z_2}} \\
{\red{x_3}} & {\red{y_3}} & {\red{z_3}} \\
{\red{x_4}} & {\red{y_4}} & {\red{z_4}} \\
\end{bmatrix}
\begin{bmatrix}
{\blue{x_1}} & {\blue{y_1}} & {\blue{z_1}} \\
{\blue{x_2}} & {\blue{y_2}} & {\blue{z_2}} \\
\end{bmatrix}
 \right ]
\rightarrow
\small 
\begin{bmatrix}
\begin{bmatrix}
{\red{x_1}} & {\red{y_1}} & {\red{z_1}} \\
{\red{x_2}} & {\red{y_2}} & {\red{z_2}} \\
{\red{x_3}} & {\red{y_3}} & {\red{z_3}} \\
{\red{x_4}} & {\red{y_4}} & {\red{z_4}} \\
{\blue{x_1}} & {\blue{y_1}} & {\blue{z_1}} \\
{\blue{x_2}} & {\blue{y_2}} & {\blue{z_2}} \\
\end{bmatrix}
\begin{bmatrix}
{\red{0}} \\
{\red{0}} \\
{\red{0}} \\
{\red{0}} \\
{\blue{1}} \\
{\blue{1}} \\
\end{bmatrix}
\end{bmatrix}
\]
\textbf{Neighborhood Aggregation}
\begin{itemize}
  \item \textbf{MESSAGE PASSING FORMAT}: Generalizing the convolutional operator to irregular domains is typically expressed as a neighborhood aggregation or message passing scheme (Gilmer et al., 2017). In the context of pointcloud, a graph is created between the support points and their neighborhood. This operation is commonly done using knn or radius search algorithms.
  Given N supports points and their respective neighborhood containing $m_i$ points, message passing scheme is leveraging gather and scatter operations based on
edge indices $ E ^ {2, \sum_{i=1}^{N} m_i}$, hence alternating between node parallel space and edge parallel space. Message passing operation doesn't introduce shadow points, but is slower than the dense operation. However, Pytorch Geometric 1.5.0 significantly improved scatter operation speed by introducing segment scattering, node ordering while finding neighbors. 

  \item \textbf{PARTIAL DENSE FORMAT}: KPconv implementation (Hugues Thomas and al.) introduced a new format where the neighborhood was padded with shadow points, so that each support points have the same number of neighbors. This format provides a great compromise between speed and sample size flexibility.  
  
  \item \textbf{SPARSE FORMAT}: With regular 3x3 convolutions, the set of active (non-zero) sites grows rapidly, which makes normal Conv3d not suitable for large pointclouds. With the introduction of Submanifold Sparse Convolutional Networks or SparseConv in short [], the set of active sites is unchanged. Active sites look at their active neighbors, and non-active sites have no computational overhead. Disconnected components don't communicate at first, although they will merge due to the effect of strided operations, either pooling or convolutions.
  Currently, three open source projects are implementing \href{https://github.com/facebookresearch/SparseConvNet}{SparseConv},
  \href{https://github.com/StanfordVL/MinkowskiEngine}{Minkowski Engine} and \href{https://github.com/traveller59/spconv}{Spconv}. Commonly used models as ResNet [], DenseNet [] or others can easily be implemented by replacing normal convolution by their respective Sparse version.
  Within those libraries, coordinates lives on cpu and features on gpu. Generating indices for convolution is done on cpu,  making training and inference slow. However, active work is undergoing to move those operation on GPU using SlabHash [].

\paragraph{SUPERPOINT FORMAT}

Points are clustered into geometrically simple yet meaningful shapes, called SuperPoints. Within Super Point Graph [], a graph is being built using knn, then l0-cut pursuit algorithm introduced by [] is applied to perform graph-cut into meaningful SuperPoints. SPG + SSP [] introduced supervized learning framework for oversegmenting 3D point clouds into superpoints. This format allows massive pointcloud and is well suited for sparse outdoor data to be processed as only a few points are sampled within each superpoint.
  
\end{itemize}

The framework support both "dense" and "heterogeneous" format on CPU / GPU for methods such as radius search, grid sampling, interpolation and many more.

\fi
\section{Numerical Experiments}
{In this section, we present several case studies demonstrating some of the capabilities of \TPD~, such as backbone swapping and fair benchmarking. The detailed configuration of the experiments, along with the evolution of metrics along the runs, are available as WandB projects accessible through the framework GitHub repository.}
\subsection{VoteNet with Different Backbones}
The VoteNet network, introduced by Qi \etal \cite{qi2019deep}, performs end-to-end object detection in 3D point clouds. It relies on a PointNet++-like backbone network to extract point features, which are then used by an object-center voting module and a box-proposal module.

In \tabref{tab:votenet}, we assess the performance of different networks  by replacing the PointNet++ backbone with more recent alternatives, such as RS-CNN \cite{liu2019rscnn}, KPConv \cite{thomas2019kpconv}, and Minkowski Engine \cite{choy20194d}.
We used the same architecture than the ones used in our semantic segmentation benchmark, see \secref{sec:benchmark}. Further adaptation to the task, such as truncating the decoder as recommend in VoteNet, would certainly be beneficial but are out of the scope of this paper. Otherwise, we use the hyper parameters, training and data augmentation procedures proposed in the original work. Changing the backbone is as simple as editing the model's configuration file as presented in \coderef{list:votenet}.

While the RS-CNN and KPConv backbones underperformed, the mean average precision \@50\% IoU is improved slightly by switching to a Minkowski Engine network. Overall, the PointNet++ architecture seems particularly well-suited to the task of object detection, as also observed by Xi \etal \cite{xie2020pointcontrast}.

Note that our results differ slightly from the original paper, as our metric implementation is slightly altered: successful box recoveries only count as positive for a given class if the predicted class is correctly inferred as well.
\begin{listing}
\caption{Model configuration used for switching the backbone of VoteNet \cite{qi2019deep} to KPConv \cite{thomas2019kpconv}.}
\label{list:votenet}
\begin{python}
VoteNetKPConv:
    class: votenet2.VoteNet2
    conv_type: "PARTIAL_DENSE"
    define_constants:
        num_proposal: 256 # num. box proposals
        num_classes: 18 # semantic classes
    backbone:
        model_type: "KPConv" # backbone type 
        extra_options:
            in_grid_size: 0.05 # input grid size
\end{python}
\end{listing}
\begin{table}
\caption{Impact of the backbone choice on VoteNet performances. mAP@$r$ stands for the interclass mean average precision with a detection threshold of $r\%$ IoU.}
\label{tab:votenet}
\begin{center}
\begin{tabular}{ccccc}
\multirow{2}{*}{\begin{tabular}{c}\textbf{VoteNet}\\ \textbf{Backbone}\end{tabular}}
& mAP& mAP \\ 
& @25 & @50\\ \hline
PointNet ++ \cite{QiYSG17PointNetPP}  & \bf{54.2} & 30.1 \\
RS-CNN \cite{liu2019rscnn} & 51.6 & 29.5 \\
KPConv \cite{thomas2019kpconv} & 48.9 & 29.2 \\
Mink. Engine \cite{choy20194d} & 53.8 & \bf{30.2}\\
\end{tabular}
\end{center}
\end{table}
\subsection{Benchmarking with \TPD}
\label{sec:benchmark}
As one of the first and easiest to use datasets, S3DIS \cite{armeni20163d} has been used as a standard measure of the performance of state-of-the-art methods. However, there is a large discrepancy in how methods are evaluated, which makes it hard to compare their performance.
We propose a common evaluation protocol, largely inspired by the one proposed by  \cite{thomas2019kpconv}.
\begin{itemize}
\item \textbf{Pre-processing:} S3DIS is comprised of $6$ folds, each containing a collection of point clouds corresponding to single rooms. We aggregate these clouds to obtain one cloud per fold, each corresponding to one entire level of an office building. We sample this cloud with respect to a $4$cm grid.
\item \textbf{Training:} In each epoch, we sample $3\,000$ spheres of radius $2$m, centered around random points picked with a probability inversely proportional to the square root of their class frequency.
\item \textbf{Optimizers:} The parameters of the optimizers are given in the configuration file presented in \coderef{list:optim}. In our experiments, Stochastic gradient Descent (SGD) had slower convergence but higher generalization than momentum-based optimizers such as \cite{kingma2014adam}.
\item \textbf{Inference:} $2$m-radius spheres are sampled along a $2\times2\times2$m grid \emph{once}. The class probability associated to points present in several spheres are averaged. To compute the final metrics on the original clouds, we perform nearest neighbor interpolation. 
\item \textbf{Metrics:} We report the Overall Accuracy (OA) and Mean Intersection over Union (mIoU) over classes obtained by cross-validating over the $6$-folds. We add an early stopping scheme in which the model is evaluated on the epoch whose model has the highest mIoU on a validation set, obtained by withholding selected rooms from the training set. 
\end{itemize}
We also devise a similar protocol for ScanNet \cite{dai2017scannet}. The differing steps are as follows:
\begin{itemize}
\item \textbf{Pre-processing:} We subsample with a $5$cm grid.
\item \textbf{Training:} Batches are collections of rooms subsampled to $50\,000$ points. 
\item \textbf{Metrics:} We report the OA and mIoU on the validation set using the model of the last epoch. 
\end{itemize}
%
%
\begin{listing}
\caption{Optimizer hyper-parameters used for training all models on S3DIS.}
\label{list:optim}
\begin{python}
epochs: 300 # Number of epochs
batch_size: 8
optim:
    base_lr: 0.01 # Learning rate
    grad_clip: 100 #gradient clipping
    optimizer:
        class: SGD # Optimizer
        params: # SGD parameters
            momentum: 0.02 
            lr: \${training.optim.base_lr}
            weight_decay: 1e-3
    lr_scheduler:
      class: ExponentialLR
      params:
        gamma: 0.9885 # /10 every 200 ep.
    bn_scheduler: # Batch Normalization Scheduler
        bn_policy: "step_decay"
        params:
            bn_momentum: 0.02 
\end{python}
\end{listing}
In \tabref{tab:bench}, we report the performance of four networks (PointNet++ \cite{QiYSG17PointNetPP}, RS-CNN \cite{liu2019rscnn}, Minkowski Engine \cite{choy20194d}, and KPConv \cite{thomas2019kpconv}) on both S3DIS and ScanNet. Each of these algorithms share the exact same learning and inference procedure. This allows us to appreciate their performances \emph{all other things being equal}.

Ze Liu \etal \cite{liu2020closerlook3d} interestingly demonstrate in their recent investigation that the choice of convolution type (pointnet-like, pointCNN \cite{li2018pointcnn}, KPConv, and their own PoolPos) have little impact when evaluated with a shared architecture. We can hence attribute the performance in \tabref{tab:bench} to the difference in architectures, namely the depth and subsampling/upsampling operations. We also remark that Ze Liu \etal's experiments are particularly easy to replicate with \TPD{} on any of the proposed dataset, as most convolution schemes are already implemented.
%
\begin{table}
\caption{Benchmarking of four different methods on the task of semantic segmentation for two different datasets: S3DIS \cite{armeni20163d} with 6-fold cross validation and ScanNet \cite{shapenet2015} evaluated on the valdiation set.}
\label{tab:bench}
\resizebox{\columnwidth}{!}{
\begin{tabular}{ccccc}
\multirow{2}{*}{\textbf{Model}}
& \multicolumn{2}{c}{ S3DIS 6-folds} 
& \multicolumn{2}{c}{ScanNet} 
\\\cmidrule(l{5pt}r{5pt}){2-3}\cmidrule(l{5pt}r{5pt}){4-5}
 &OA&mIoU&OA&mIoU\\\hline
 KPConv \cite{thomas2019kpconv}& \bf86.4 & \bf66.3 & 85.5 & 59.9 \\
 Mink. Engine \cite{choy20194d}& 86.0 & 65.9  & \bf87.2 & \bf65.0\\
 RS-CNN \cite{liu2019rscnn} & 83.2 & 62.9 & 79.8 & 47.2\\
 PointNet++ \cite{QiYSG17PointNetPP}& 81.06 & 56.7 & 80.7 & 49.3 \\
\end{tabular}
 }
\end{table}
%
\subsection{Inference Schemes}
A common scheme to increase the performance of a model is to perform several inference runs---with data augmentation, and to output their average probability. While this method slows down inference, the increase in performance can be justified for non time-sensitive applications such as digital twin modeling.

\begin{figure}[]
\centerline{\includegraphics[scale=0.55 ]{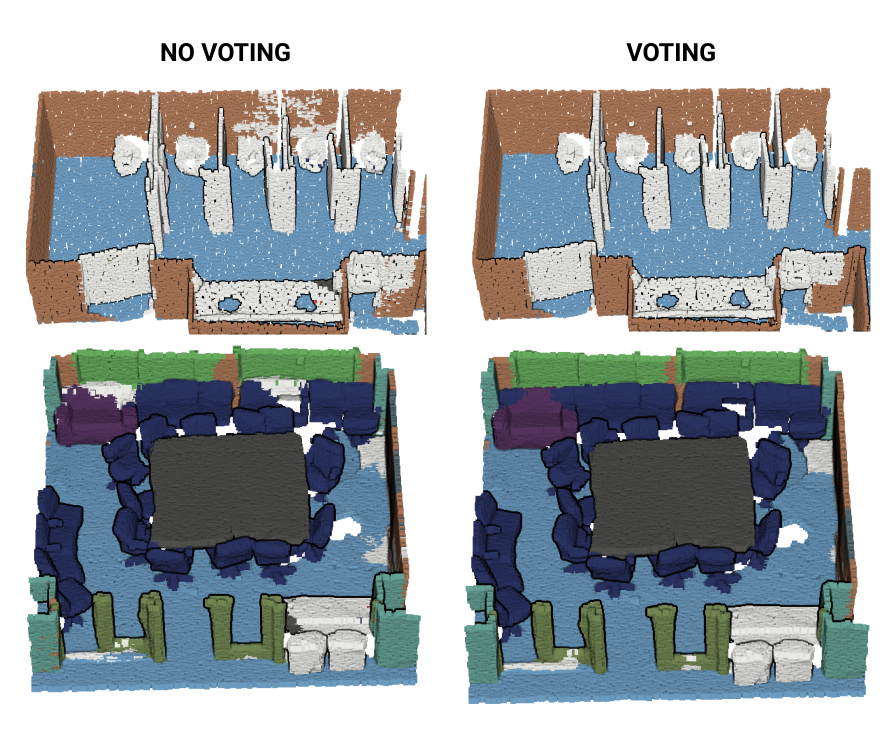}}
\caption{Segmentation predictions with and without voting. We can observe voting tends to create smoother, more accurate predictions.}
\label{fig:augment}
\end{figure}

In \tabref{tab:augment}, we report the performance of different models with and without a $3$-run average. The performance increase is noticeable, with an average increase from $1$ to $3$ points of mIoU. Interestingly, we remark that the relative order of performance of KPConv and Minkowski Engine are reversed by a non-negligible margin when using this inference scheme. In \figref{fig:augment}, we illustrate the improvement provided by this inference scheme.

\begin{table}
\caption{Improvement in terms of mIoU provided by inference-time voting schemes on S3DIS 6-folds.}
\label{tab:augment}
\centering
\begin{tabular}{ccccccc}
 Models& no voting &with voting\\\hline
 KPConv \cite{thomas2019kpconv} & \bf66.3 & 67.2 \\
 Mink. Engine \cite{choy20194d} & 65.9 & \bf69.1 \\
  RS-CNN \cite{liu2019rscnn}& 62.9 & 64.6\\
PointNet++ \cite{QiYSG17PointNetPP}& 56.7 & 59.0  \\
\end{tabular}
\end{table}
\subsection{Registration}
Registration is the task of finding a rigid transformation aligning several 3D point clouds. Neural networks can be trained to compute point features whose pairing determine the sought-after transformation, either end-to-end or with robust nondifferentiable estimators such as RANSAC, FGR\cite{leibe_fast_2016}, or the recent TEASER\cite{yang2020teaser}.

\begin{figure}
    \centering
    \begin{tabular}{c}
    \begin{subfigure}{0.9\columnwidth}
    \includegraphics[width=1\columnwidth]{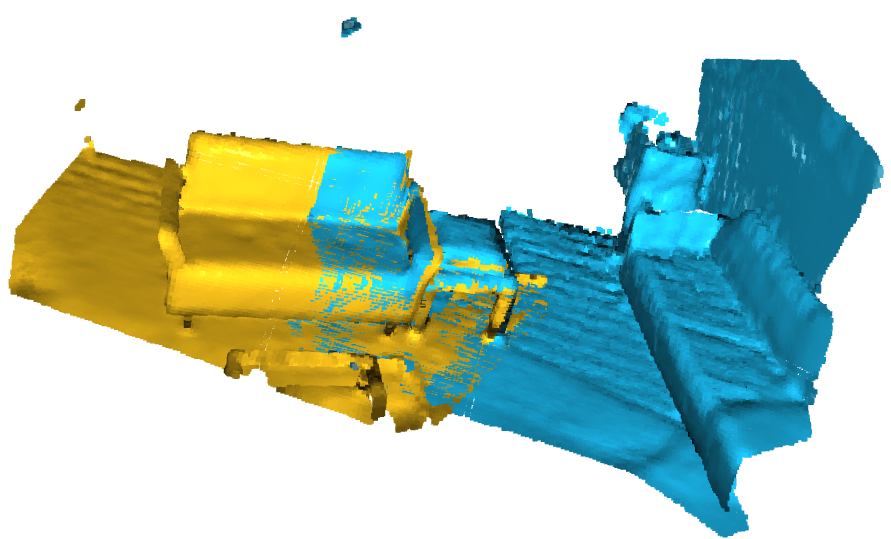}
    \caption{Registration example on 3DMatch.}
    \label{fig:3dm}
    \end{subfigure}
    \\
     \begin{subfigure}{0.9\columnwidth}
    \includegraphics[width=1\columnwidth]{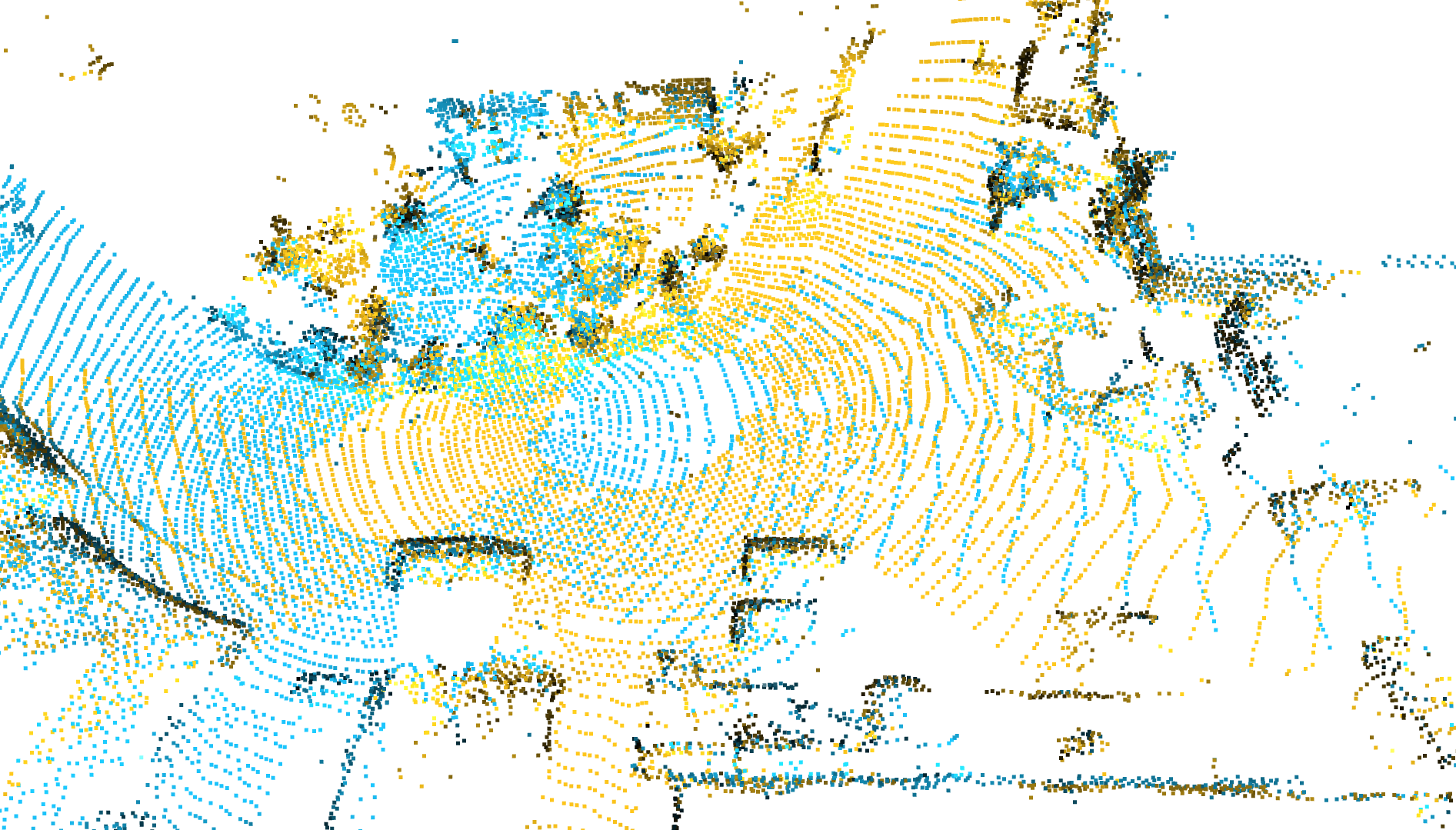}
    \caption{Registration example on KITTI odometry.}
    \label{fig:kitti}
    \end{subfigure}
    \end{tabular}
    \caption{Qualitative registration results between two point clouds (in blue and yellow).}
    \label{fig:registration}
\end{figure}

We implement a full registration pipeline within \TPD{}, using a Minkowski Engine backbone as suggested by Choy \etal \cite{choy2019fully}, and estimating transformations with TEASER\cite{yang2020teaser} and RANSAC. As reported in \tabref{tab:registration}, our implementation reaches state-of-the-art performance of point cloud registration on two datasets available in \TPD{}:  3DMatch \cite{zeng20163dmatch} and KITTI Odometry \cite{Geiger2012CVPR}. In \figref{fig:registration}, we present some qualitative illustrations.
\begin{table}[]
		\caption{Success rate (in \%) on the 3DMatch and KITTI odometry datasets. Results obtained within TorchPoints3D.
		A \emph{success} is defined by an error under $15$ degree and $0.3$ m for 3DMatch, and $2$ degrees and $0.6$ m for Kitti Odometry.
		$^\star$ model taken from the paper's repository.}
		\label{tab:registration}
		\centering
		\begin{tabular}{ccc}
			Methods & 3DMatch & KITTI \\ \hline	
		    DGR$^\star$  & 92.6 & 96.9 \\
		    FCGF
		    + TEASER  & {93.6} & {\bf 99.8} \\
		    FCGF
		    + RANSAC & {\bf 94.3} & {99.6}
		\end{tabular}%
\end{table}
\subsection{CPU-Based Preprocessing}
When training point-based neural networks on large point clouds, the computation of neighbors and the sub-sampling operations become the computational bottlenecks rather than inference or backpropagation.

In modern deep learning frameworks such as PyTorch, background processes prepare new batches of data to be run through the network while the GPU simultaneously performs tensor operations on previously prepared batches. As suggested in \cite{thomas2019kpconv}, we off-load the radius search and sub-sampling operations to those background processes operating on CPUs. As reported in \tabref{tab:speed}, this allows us to achieve an $8$-times overall speed-up compared to performing all computations on the GPU. This particular implementation trick---one of many---exemplifies the numerous pitfalls to overcome when implementing deep learning methods operating on 3D data.

\begin{table}[!h]
\caption{Training speed of the KPconv model, in thousands of points processed per second (kpts/s) during training, with radius search performed on either the GPU (Tesla T4) or the CPUs ($4\times 2.2$GHz).}
\label{tab:speed}
\centering
\begin{tabular}{ccc}
                &  S3DIS & ScanNet\\ \hline
   radius search on CPUs & \bf$199.9$ & \bf$197.4$ \\
   radius search on GPU & $38.6$ & $29.2$
\end{tabular}
\end{table}

\section{Conclusion and Perspectives}
We presented \TPD, a flexible and powerful framework aiming to make deep learning on 3D data both more accessible and rigorous. Our implementation allows users to evaluate, improve and combine state-of-the-art models on a growing number of tasks and datasets. 
The community emerging around our framework provides us with precious feedback, as well as much needed help in keeping up with a such a fast-paced domain. We welcome researchers, software engineers, and open-source enthusiasts in this endeavor.

Encouraged by the recent results of Xie \etal\cite{xie2020pointcontrast}, we believe in the potential of transfer learning across datasets and tasks for 3D data. Our next focus will be to provide a high-level API for pre-trained, self-supervised, self-trained, and unsupervised deep learning approaches operating on 3D point clouds.

\section{Acknowledgments}
Thomas Chaton and Nicolas Chaulet would like to acknowledge the great support provided by Fujitsu Laboratories of Europe, both financially and for useful feedback by being one of the first users.\\
This research was also supported in parts by the AI4GEO project: \url{http://www.ai4geo.eu/}

\pagebreak

{\small
\bibliographystyle{ieee}
\balance
\bibliography{biblio.bib}
}

\newpage
\section*{APPENDIX}
\ARXIV{\appendix
\setcounter{section}{0}

\section{Illustrative Notebooks}

We propose two IPython notebooks to illustrate some of the capacities of \TPD:
\begin{itemize}
    \item A notebook illustrating the training of the Relation-Shape CNN model for object classification with \TPD{}:  \href{https://colab.research.google.com/drive/10Pryg73xoJzkkBuZB2pFgTbwLruO_0Za?usp=sharing}{link}.
    \item A notebook showing the inner working of encoding and decoding in a KPConv model for part segmentation:  \href{https://colab.research.google.com/drive/1wpPESfw7bSrcN-AE52Dr_V5b2ps90MHt?usp=sharing}{link}.
\end{itemize}
These notebook can be run on Google Colab from a browser without any installations. They are self-contained and will install automatically all required packages, as well as download the relevant datasets. Be warned that the installation of the necessary libraries and download of the datasets can take up to $30$minutes. You can otherwise download the notebooks and run them locally, after installing the necessary libraries, namely torch torch-points3d and pyvista.
\begin{listing}
\label{code:shapenet}
\begin{python}
pip install torch torch-points3d pyvista
\end{python}
\end{listing}
%
%
\section{Details on the Registration Experiment}
\subsection{Implementation}
As described in the main paper, we implemented FCGF model for point cloud registration. 
The encoder is composed of 4 residual blocks with output sizes of $[32, 64, 128, 256$]. Each block has a stride of $2$ except for the first block. The decoder is composed of residual blocks of output sizes $[64, 64, 64, 64]$. We train with SGD with momentum of $0.8$, a learning rate of $0.1$, and a weight decay of $10^{-4}$. The same parameters are used for both Kitti Odometry and 3DMatch. 
\subsection{3Dmatch}
Since 3D match is a dataset of RGBD images, we need to fuse depth images to obtain 3D point clouds. We use a TSDF voxel grid as in \cite{CurlessL96} and obtain fragments. For the training set and the validation set, $50$ depth images are fused to obtain each fragment.
We use a voxel subsampling of size $0.02$m in the FCGF network.  
\subsection{KITTI Odometry}
Kitti Odometry contains LiDAR scans with their associated poses, however these do not have the precision necessary to properly evaluate registration predictions. Choy \etal \cite{choy2019fully} use these poses as  initialization, and use ICP to refine the transformation between each pairs. The pairs are defined as LiDAR frames whose center is at least distant of $10$m. We use IMLS SLAM \cite{deschaud2018imls} to compute the poses of all sequences. 
As in \cite{choy2019fully, choy2020deep}, the sequences $0, 1, 2, 3, 4, 5$ are used for training, $6, 7$ for validation and $8, 9, 10$ for testing.
For this dataset, we use a voxel subsampling of size $0.3$m.
\subsection{Evaluation}
To evaluate registration, we measure the rotation error and the translation error as :
\begin{align}
    \delta_{trans} &= ||t_{est}-t^\star||_2\\
    \delta_{rot} &= \arccos{\frac{trace(R_{est}R^{\star\;T})-1}{2}}
\end{align}
where $(t_{est}, R_{est})$ is the estimated translation and rotation, and
$(t^\star, R^\star)$ the associated ground truth.
For 3DMatch, we count a \emph{success}  when the rotation error is under $15$ degrees, and the translation error is less than $0.3$ m, as suggested by Choy \etal  \cite{choy2020deep}
For Kitti Odometry, the rotation error must be less than $2$ degrees and the translation error must be under $0.6$ m for a prediction to be considered successful.
\begin{figure}[h]
    \centering
    \includegraphics[width=\columnwidth]{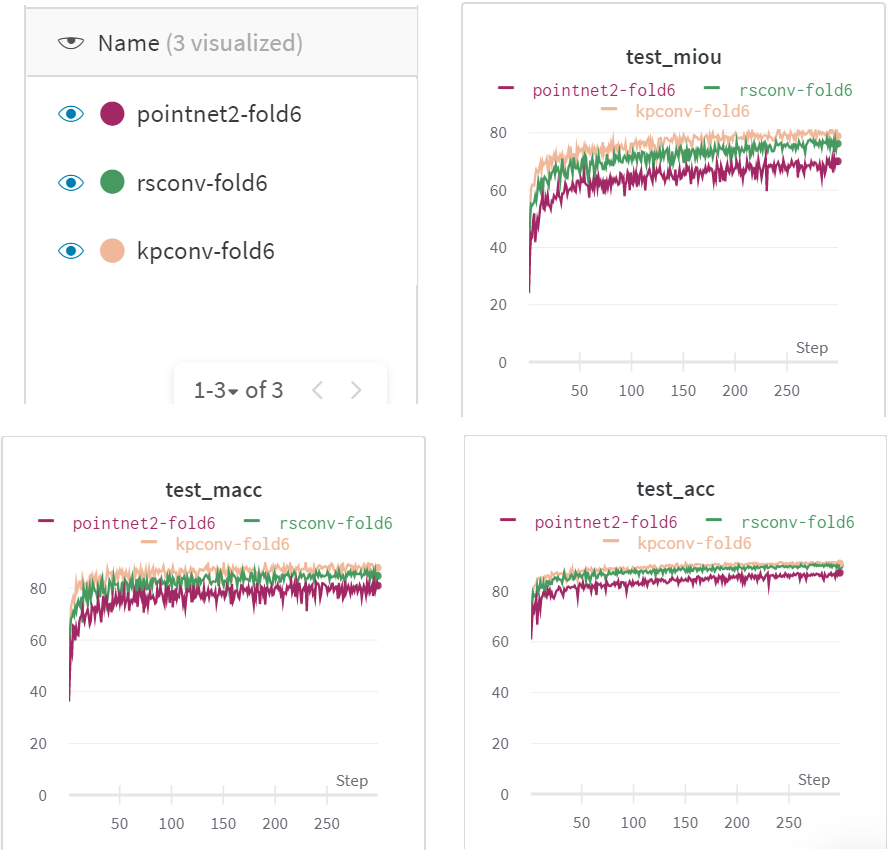}
    \caption{Training logs hosted on \url{wandb.ai} for different models on a S3DIS fold. These logs, along their trained models and full configuration, are publicly available on the contributor's  \url{wandb.ai}'s account (links provided on the repository).}
    \label{fig:wandb}
\end{figure}}{}
\end{document}